\RequirePackage{fix-cm}
\documentclass[twocolumn]{svjour3}          
\smartqed  
\usepackage{graphicx}
\usepackage{caption}
\usepackage{subcaption}

\usepackage[table]{xcolor}
\usepackage{caption}

\usepackage[intlimits]{amsmath}     
\usepackage{amsthm}
\usepackage{amssymb}
\usepackage{nomencl}
\usepackage{acronym}
\usepackage{color}

\begin{document}

\title{
 Cooking in the kitchen: Recognizing and Segmenting Human Activities in Videos  \\
}

\titlerunning{Cooking in the kitchen: Recognizing and Segmenting Human Activities in Videos}        

\author{Hilde Kuehne         \and
        Juergen Gall         \and
        Thomas Serre 
}


\institute{Hilde Kuehne \at
              University of Bonn\\
              \email{kuehne@iai.uni-bonn.de}       
           \and
           Juergen Gall \at
              University of Bonn\\
              \email{gall@iai.uni-bonn.de}           
           \and
           Thomas Serre \at
              Brown University \\
              \email{thomas\_serre@brown.edu}           
}

\date{Received: date / Accepted: date}


\maketitle

\begin{abstract}

As the field of action recognition matures, research is rapidly moving away from simpler problems such as action recognition in short hand segmented video segments to more complex real-world problems such as the continuous monitoring and analysis of daily human activities. 

We propose an end-to-end generative approach for the segmentation and parsing of complex human activities. In this approach, a visual representation based on reduced Fisher Vectors is combined with a structured generative temporal model for recognition. To overcome one of the major limitations of generative models, i.e., their need for large amount of training data, we recorded a large scale activity dataset featuring 52 participants preparing 10 distinct dishes in their own kitchen. We annotated the resulting video at both a coarse and fine level of action granularity. The dataset was used to evaluate the proposed approach on various tasks ranging from basic action unit classification to activity recognition as well as the segmentation and parsing of video sequences. Our results demonstrate the ability of structured temporal generative approaches to cope with the complexity of daily-life activities. 

\keywords{Activity classification, Action detection, Video segmentation, Action recognition}
\end{abstract}

\section{Introduction}
\label{sec:intro}

Several real-world applications of computer vision including smart homes, surveillance and assisted living require continuous video monitoring. However, to date, most of the work on action recognition still focuses on the somewhat simpler problem of assigning class labels to short pre-segmented video clips. In comparison, methods for the automated analysis of temporal structures, including methods for parsing and segmentation are still in their early stages of development. Progress in action recognition has been largely spurred by the increasing availability of large realistic video datasets that allow the benchmarking of different visual representations and classification methodologies. Unfortunately, there is currently a dire need for similar large-scale datasets comprising long video sequences of complex activities recorded ``in the wild''. 

To fill in this void, we describe a novel human-activity video dataset that we named the Breakfast dataset. This dataset includes almost 70 hours of hand-annotated videos corresponding to 52 unique participants preparing 10 distinct breakfast dishes in 18 different home kitchens. The dataset provides annotations at two different temporal scales, a finer scale, corresponding to low-level task-oriented motion sequences such as ``open drawer'' - ``reach knife'' - ``carry knife'', and a coarser scale corresponding to higher-level goal-oriented action sequences, such as  ``take plate'' - ``cut fruit'' etc.  The proposed dataset is currently the largest available data\-set for human activity recognition and will allow for benchmarking of  various activity recognition approaches as well as the parsing, segmentation and  detection of activities into finer action units.
In addition to supporting general research in computer vision, we hope that our dataset will contribute the testing of brain theories of event perception. A  body of the cognitive psychology literature is devoted to understanding how people perceive human actions over time and, in particular, what brain mechanisms support the perception of human movements. In this context, it has been shown that activities are not perceived as a continuous input stream but rather as discrete action units within a behavioral sequence~\cite{Baird2001Making}. The segmentation of action streams and the detection of discrete boundaries between action units happens at different levels of granularity such that action segments get combined over time to form a holistic interpretation of perceived actions~\cite{Zacks2007Event}. This abstraction has also been shown to be a condition for people to predict and react to others' intentions (see, e.g., ~\cite{Zacks2009Using}), and can thus be seen as a fundamental ability towards our understanding of  going activities. 
We thus hope that the release of a large activity dataset together with consistent behavioral annotations at both fine and coarse levels of granularity combined with visual representations derived from state of the art computer vision systems will help further our understanding of the brain mechanisms underlying event perception.
%

We further describe a novel generative framework for the analysis of temporal structures. In this framework, Fisher Vectors (FVs) are used to represent individual video frames and action units are modeled by Hidden Markov Models (HMMs). Action units are combined through an activity grammar that is learned from data. We extensively evaluate the resulting approach on the proposed Breakfast dataset as well as four other benchmarks. Our evaluation, which includes activity recognition, action unit classification and segmentation, demonstrates that the approach performs on par or better than the state-of-the-art.       

A preliminary version of this work appeared in~\cite{Kuehne2014Language,Kuehne16}. The present work extends our original release of the Breakfast dataset~\cite{Kuehne2014Language} with additional fine-grained annotations. We also provide a more comprehensive overview of the generative framework used to analyze temporal structures first presented in~\cite{Kuehne16}. The present evaluation has also been extended compared to that in~\cite{Kuehne16} to include action unit classification and segmentation at different levels of granularity, activity recognition as well as a complete run-time analysis of the system and its dependency on the amount of training data available. 

\section{Related Work}
\label{sec:relatedWork}

Before discussing structured temporal models and activity recognition datasets in Sections~\ref{subsec:rw_temp_aspects} and~\ref{sec:relatedWork:datasets}, we first briefly review the state-of-the-art in video-clip classification~\ref{sec:relatedWork:AC}.

\subsection{Unstructured approaches to video-clip classification}
\label{sec:relatedWork:AC}
The best current approaches to human action recognition in video clips rely on dense trajectory features~\cite{Wang2013Dense} that are then quantized using the Fisher Vector (FV) method. The combination of Fisher vector encoding and dense trajectories was first described in \cite{Wang2013Action,Oneata2013Action} and shown to achieve state-of-the-art classification accuracy on several action datasets. The approach was further improved in~\cite{Peng15Action} using stacked FVs. In addition to FVs, it was shown that the accuracy of this approach could be improved by modelling the context of an action. This was done in~\cite{jain201515} via the detection of objects in the scene using a convolutional neural network. More generally, deep learning networks have also been described to learn temporal features. For instance, CNNs were used for the training and classification of 1 million YouTube videos~\cite{karpathy2014large}. 
In addition, the combination of learned features derived from a CNN and hand-crafted features was shown to be promising~\cite{Wang15ActionTDD}.

\subsection{Structured temporal models in activity recognition}
\label{subsec:rw_temp_aspects}

Because of an increased interest in continuous monitoring and analysis of human activities, several structured temporal models have been recently proposed for the recognition of complex activity sequences.

Early approaches were based on motion-captured data~\cite{Guerra2005Discovering,Sminchisescu2005Conditional,Lillo2014Discriminative} or hand-labeled trajectories~\cite{Rao2002ViewInvariant}. One of the first attempts to generalize these methods to raw video data was presented in~\cite{Niebles2010Modeling}. In their approach, the authors proposed to classify human activities by aggregating information from motion segments based on visual features and temporal compositions. Video sequences were thus decomposed into temporal segments of variable length and matched against motion segment classifiers.  The idea of representing a video by snippets was later adapted in diverse forms including Actoms~\cite{gaidon2011Actom}, action spectograms~\cite{ChiaChih2011Modeling}, middle-level components~\cite{Yuan2012MiddleLevel} or clusters of Tracklets~\cite{gaidon2014Activity}.

The development of approaches for the recognition of complex events has also gained in popularity. Early approaches modeled temporal structures using velocity history models~\cite{Messing2009Activity}, Bayes networks~\cite{Ryoo2009Human} or hybrid HMMs~\cite{ChiaChih2011Modeling}. However, the datasets used for evaluating these approaches tended to be relatively simple. With the increased complexity of available datasets, the visual representations used by modern approaches has also become increasingly complex. The temporal dynamics of video sequences was modeled in~\cite{Bhattacharya2014Recognition} using vector time series represented by the principal projections of an eigenvector decomposition of their block Hankel Matrix and harmonic signatures. 
The resulting mid-level representations were successfully applied to the recognition of complex events using the TRECVID dataset. 
In~\cite{Cheng2014Temporal}, the authors used a sequence memorizer~\cite{Wood09Stochastic}, i.e., a hierarchical nonparametric Bayesian model that captures long-term dependencies in sequence data. 
A higher level representation based on a stochastic context-free grammar was proposed in~\cite{Vo2014Stochastic}. 
Another approach for constructing an activity grammar automatically to capture hierarchical temporal structures was proposed in~\cite{Pirsiavash2014Parsing}. In this approach, parsing was based on a latent structural SVM which learns sub-actions automatically. A similarly unsupervised method for learning action units was described in~\cite{wu2015watch}. Here, a causal topic model was used to learn the co-occurrence and temporal relation between action units in videos.
Another system was described in~\cite{Soran15Generating} with the goal to detect missing actions within an activity sequence. The system combines coupled HMMs with a higher-level graph to model the overall structure of an activity. Based on the detection of omitted nodes/action units within the graph, the system produces notifications for missing action units.

\subsection{Datasets}
\label{sec:relatedWork:datasets}

\begin{table*}[t]
   \centering
\begin{tabular}{|c|c|c|c|c|c|c|c|c|c|c|c|}
\hline 
         & Activities & Units  &  Segments  &  Clips &  Duration  & Setting  & Persons & Mean length  \\
\hline 
 Cha LAP        & -   & 11          &     166     &  7     &    6min     &   staged     &   8     &  2.1sec \\
\hline  
 YouCook        &  -  &    7        &     1422     &   88(46)     &   0.7h     &    youtube   &    -     &  1.9sec  \\
\hline 
 Toy assembly   &  3  &  40         &    479    &   29    &   1.06h   &     lab    &    2     &  6.4sec \\
\hline    
 CMU MMAC (*)   & 1   &  37      &     2238     &   37      &  4.4h    &     lab     &    16    &  6.8sec \\
\hline         
 50 Salads      & 2   &  51      &  2603     &  50    &   5.3h     &     lab     &   25    &  7.0sec \\
\hline        
 MPII Cooking   & 14  &  65         &      5609 (1861 BG)      &   44     &   8.1h    &     lab    &   12    &  8.0sec  \\
\hline 
 MPII Composite &  55  &  78        &   12642    &   256    &   18.2h   &     lab    &     30    &   5.1sec  \\
\hline 
\hline 
 Breakfast         & 10  &  48         &   11441 (2992 BG)   &  1712   &   66.7h   &    wild   &     52    &  26.0sec \\
\hline        
 Breakfast Fine    & 10  &  178         &  31325 (2154 BG)   &  804   &    49.6h   &   wild    &     52    &  5.6sec  \\
\hline 
\end{tabular} 
\caption{Overview of existing datasets available for video segmentation evaluation and comparison with the proposed Breakfast dataset. Note that we only consider videos with action segment annotations, which, in some cases (e.g., YouCook or CMU MMAC) correspond to only a subset of the entire dataset. If background class labels are included in the annotations, the corresponding number of segments is included in parenthesis.}
\label{tab:overview_datasets}
\end{table*} 

With the development of novel structured temporal models, various datasets have been described to evaluate the temporal parsing of activities. As there are many datasets for (clip-based) action classification (for an overview see e.g.~\cite{Chaquet13survey}), we here only focus on video datasets that provide temporal segment annotations with at least one level of granularity. Some of the most recent activity recognition datasets that provide such annotations include the CMU MMAC dataset~\cite{Spriggs2009TemporalSegmentation},  the MPII Cooking dataset~\cite{rohrbach12cvprOnline} and the 50 Salads dataset~\cite{Stein2013Combining}.

Most of these datasets are mid-size, they comprise only few subjects, and the number of clips per class tend to be small. Much like those available for action classification datasets such as HMDB~\cite{Kuehne11HMDB} or UCF101~\cite{Soomro12UCF101}, which comprise 50--100 classes with 100 or more clips per class, there is a need for large-scale datasets for the recognition of complex everyday activity sequences. While this void is partially filled by the MPII Composite dataset~\cite{Rohrbach2012Script}, the dataset remains limited in that it was recorded in a lab environment with a fixed camera setup and constant lighting conditions. In contrast to the present Breakfast dataset which was recorded ``in the wild'' using 18 real-world kitchens with uncontrolled camera positions and uncontrolled light conditions. 
Additionally, the behavior of staged participants in a lab setting may differ from their everyday behavior at home. 
Indeed, recent action classification benchmarks have demonstrated the importance of unconstrained settings: Most existing systems achieve near ceiling accuracy on staged datasets such as KTH~\cite{Schuldt04Recognizing} or Weizmann~\cite{Blank05Actions} but, in comparison, very few exhibit high accuracy on datasets collected in the wild such as HMDB or UCF101. 


A second limitation of existing datasets is that the level of granularity of the action annotations vary from datasets to datasets.  Most datasets include annotations based on human object interactions such as ``open brownie box'' (CMU MMAC)  or ``screw open'' (MPII Cooking). But the labeling can also be based on more than one level granularity. For instance, in 50 Salads, where coarser elements such as ``preparing salad'' are made up from a set of finer action units such as ``cut tomato'' or ``peel cucumber'' and the finer actions are again partitioned into a preparation, core and post phase. 
Some datasets further provide labels for complex activity classes. Usually, one video represents one activity class.
Examples for such long term activities can be found in the CMU MMAC or the MPII Cooking dataset for example. 

Our dataset provides labels for long-term activities and segmentation at two levels of granularity. At the finest level, a segment is about 6 seconds long on average. At the coarser level, segment duration is about 26 seconds on average. See Table~\ref{tab:overview_datasets} for a complete comparison of existing datasets with the proposed Breakfast dataset.

\section{Breakfast Dataset}
\label{sec_datasets_BF}

\subsection{Breakfast data collection}

We describe the  Breakfast dataset for the evaluation of structured temporal recognition models. The dataset features 52 unique participants, each engaged in 10 distinct cooking activities captured in 18 different kitchens\footnote{\url{http://serre-lab.clps.brown.edu/resource/breakfast-actions-dataset}}. 
Overall, the dataset includes about 200 clips for each cooking activity including the preparation of coffee (n = 200 samples), orange juice (n = 187), chocolate milk (n = 224), tea (n = 223), a bowl of cereals (n = 214), fried eggs (n = 198), pancakes (n = 173), a fruit salad (n = 185), a sandwich (n = 197) and scrambled eggs (n = 188).

\begin{figure}
\centering
\includegraphics[width=0.45\textwidth]{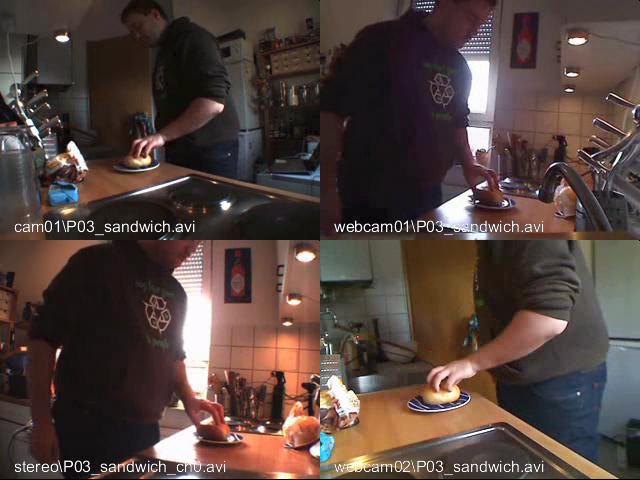} \hspace{0.01cm} 
\includegraphics[width=0.45\textwidth]{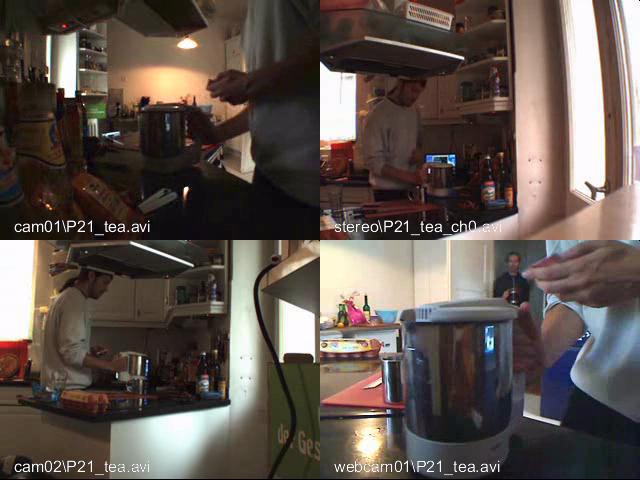}  \\ \vspace{0.1cm}
\caption[Examples from the Breakfast dataset]{Sample images from the Breakfast dataset }
\label{fig_example_bf}
\end{figure}

All activities were recorded with three to five cameras that were placed at various positions in the kitchens, so that the same activity is recorded from different, varying, viewpoints (Figure~\ref{fig_example_bf}). For the recording, webcams, standard industry cameras (Prosilica GE680C) as well as a stereo camera (BumbleBee\textregistered, Pointgrey, Inc) were used.  
All videos were normalized to a resolution of $320\times 240$ pixels with a frame rate of 15 fps. The video streams of each camera was manually synchronized. 
Overall the dataset provides about 66 hours of video and about 3.5 millions frames. 
For evaluation purpose, we organized the 52 participants in four groups, and permuted each of these four groups as splits for training and test. 

The recording setup is ``in the wild'' as opposed to a single controlled lab environment~\cite{rohrbach12cvprOnline,Spriggs2009TemporalSegmentation} in order to closely reflect real-world conditions as it pertains to the monitoring and analysis of daily activities. The actor performance was completely unscripted, unrehearsed and undirected. The actors were only handed a recipe and were instructed to prepare the corresponding food item. The resulting activities are thus highly variable both in terms of the choice of individual action units executed by the actors and their relative ordering. Since the sequences were recorded in various kitchens, the participants used the tools and packages that were locally available.
Examples of the various settings and viewpoints are shown in Figure~\ref{fig_example_bf}.

The set of  cooking activities was chosen to include many similar elements (e.g., fried egg vs.\ scrambled egg preparation, or tea vs.\ coffee) resulting in shared action units (e.g., crack egg or take cup) across activities. 
This should yield a low inter-class variance for activities combined with a high intra-class variance because of  different recoding locations, view-points and kitchens used. This challenging dataset thus allows for a thorough evaluation of structured temporal approaches.

\subsection{Data annotation}

We asked two sets of annotators to manually label videos at two different levels of granularity: One group consisting of three annotators was asked to annotate action units at a coarse level (e.g., `pour milk' or `take plate'). The start and endpoints of a segment at the coarse level was typically based on the usage of a certain tool. For instance, the coarse unit `pour milk' starts when the milk package is reached and ends when the package is released again. It  comprises all action units related to this task such as the opening or closing of the package and the pouring of the milk itself. 
Overall we identified 48 different coarse action units with about 11,000 samples in total including about 3,000 `silence' samples. Table~\ref{tab_units_overview} lists the coarse action units corresponding to individual activities.
\begin{table*}[th]
\small
\begin{tabular}{|p{2.2cm}|p{14.5cm}|}
\hline
 Coffee & \begin{small} take cup - pour coffee - pour milk - pour sugar - spoon sugar - stir coffee \end{small} \\ 
 \hline
(Chocolate) Milk & \begin{small} take cup - spoon powder - pour milk - stir milk \end{small} \\ 
\hline 
 Juice & \begin{small} take squeezer - take glass - take plate - take knife - cut orange - squeeze orange - pour juice  \end{small} \\ 
\hline
 Tea & \begin{small} take cup - add teabag - pour water - spoon sugar - pour sugar - stir tea \end{small} \\ 
\hline
 Cereals & \begin{small} take bowl - pour cereals - pour milk - stir cereals \end{small} \\ 
\hline
 Fried Egg & \begin{small} pour oil - butter pan - take egg - crack egg - fry egg - take plate - add salt and pepper - put egg onto plate \end{small} \\ 
\hline
 Pancakes & \begin{small}  take bowl - crack egg - spoon flour - pour flour -  pour milk - stir dough - pour oil - butter pan  - pour dough into pan - fry pancake - take plate - put pancake onto plate \end{small} \\ 
\hline
(Fruit) Salad & \begin{small} take plate - take knife - peel fruit - cut fruit - take bowl - put fruit to bowl - stir fruit \end{small} \\ 
\hline 
 Sandwich & \begin{small} take plate - take knife - cut bun - take butter - smear butter - take topping - add topping - put bun together \end{small} \\ 
\hline
 Scrambled Egg & \begin{small} pour oil - butter pan - take bowl - crack egg - stir egg - pour egg into pan - stir fry egg - add salt and pepper - take plate - put egg onto plate   
\end{small} \\ 
\hline
\end{tabular} 
\caption{Coarse action units for individual activities. 
}
\label{tab_units_overview}
\end{table*}

\begin{figure*}[t]
\centering
\includegraphics[width=0.9\textwidth]{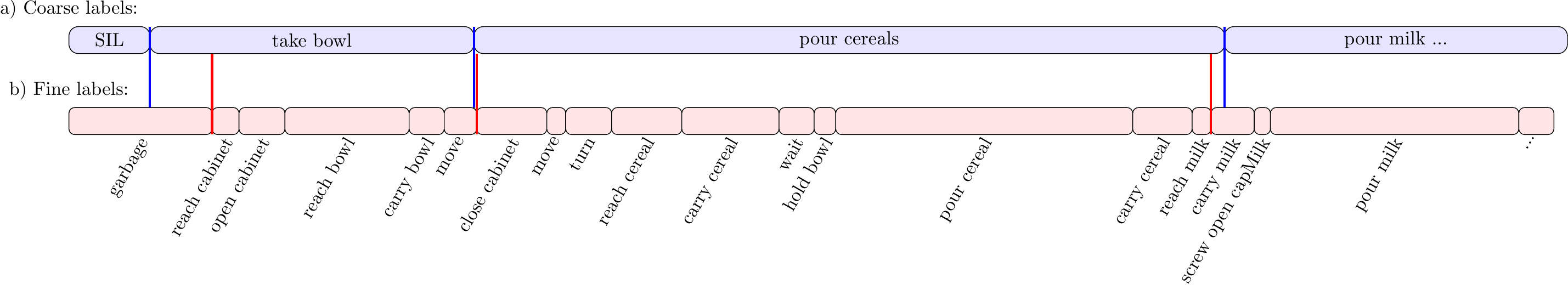}  \\ \vspace{0.1cm}
\caption{Example of coarse- and fine-level annotations for one video with the activity label ``preapre cereals''. The boundaries of the coarse units (blue) are very close to boundaries of fine unit segments (red), except for the special case of SIL/garbage. }
\label{fig_example_seg}
\end{figure*}

To address the question of how granularity influences the overall activity recognition, we asked another group of fifteen annotators to provide annotations at a finer temporal scale. At the fine level, units usually correspond to body-part movements. For instance,  a coarse unit such as `pour milk' is decomposed into finer chunks such as `grab milk' $\rightarrow$ `twist cap' $\rightarrow$ `open cap' etc.  The  mean length of the fine grained labels is 49 frames. As the task of low-level annotation is very time consuming, we only annotated a subset of 802 clips (about 28.4h of video) at the fine grained level. We refer to this subset as ``Breakfast Fine''.
 Overall, we collected about 31,000 fine grained units from 178 fine grained unit classes, including about 2,000 `silence' samples. Class labels are usually a composition of verb and object such as `reach knife', `pour water' or `cut bread'. The compositions are made up of 38 unique verbs and 62 unique objects. Note that from the 178 classes about 100 are based on the verbs `reach' or `carry' such as `reach milk' or `reach spoon'.

Although the annotators for coarse and fine grained action units worked independently, we found a high correlation between breakpoints (corresponding to transitions between action units) derived from fine- and coarse-level annotations. Examples for the segmentation of coarse and fine grained units are shown in Figure~\ref{fig_example_seg}. One can see that the boundaries of the coarse units are very close to boundaries of fine unit segments. 
To assess systematic labeling errors, we computed the frame difference of each coarse breakpoint to the nearest fine grained breakpoint. The mean distance between fine and coarse breakpoints was 12 frames and around 60\% of all coarse breakpoints were within  5 frames or less to their closest fine  breakpoint. 
On average each coarse unit comprised 6 fine-grained action units. 

\section{Representation of Activities}
\label{sec_system}

\subsection{Action unit model}
\label{sec_system:unit}


To model the temporal extent of an ongoing movement, we present an approach borrowed from speech processing. By analogy to phonemes in speech that are building blocks for words or sentences, we interpret a complex activity sequence as a concatenation of shorter action units. Given this analogy, we model action units using HMMs much like phonemes in speech processing. 

In order to model the temporal dynamics of action units, we assume that a video segment may be encoded as a sequence of feature vectors that represent the ongoing motion in each frame. The task of recognizing an action unit is therefore defined as that of finding the action unit \mbox{$u_i \in \{u_1, u_2, \dots , u_I\}$} that matches an input sequence $\textbf{x} =  (x_1, x_2, \dots , x_T)$ best, with $x_t$ representing the feature vector at frame $t$.
This can be formulated as maximizing the probability of an action unit $ u_i $ given the input sequence $\textbf{x}$:
\begin{equation}\label{equ_unit_recog}
\begin{split}
\underset{i \in 1,...,N}{\operatorname{argmax}}\:{ P(u_i | \textbf{x}) } = \underset{i \in 1,...,N}{\operatorname{argmax}}\:{ \dfrac{P(\textbf{x}|u_i)P(u_i)}{P(\textbf{x})}  } \text{ .}
\end{split}
\end{equation}
As the observation probability $ P(\textbf{x}) $  of the current sequence $\textbf{x}$ is the same for all units, it is usually omitted. The unit probability $ P(u_i) $ is in our case proportional to $ \frac{1}{N(u_i)} $ where $N(u_i)$ is the number of class samples in the training data. 

\begin{figure}
\centering
\includegraphics[width=0.99\linewidth]{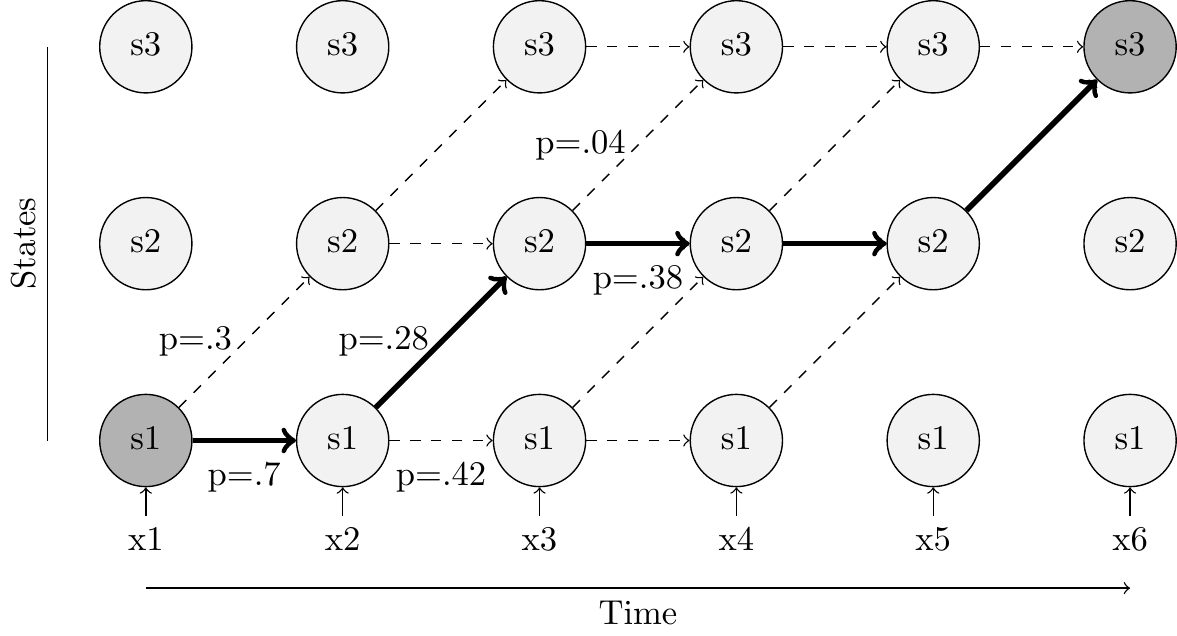} 
\caption{Inference with a left-to-right feed forward topology.  
The set of states is given by  $S = \{ s_1, s_2, s_3\} $ and the input sequence is \mbox{$\textbf{x} =  (x_1, x_2, \dots , x_6)$}, each $x_t$ corresponds to a feature vector sampled at frame $t$. The dashed lines show all possible solutions. Note that the sequence has to start with $s_1$ and end with $s_3$. Furthermore, only transitions to the next state are allowed. The path with the highest probability (bold) is obtained using the Viterbi algorithm.
}
\label{fig:HMM_graph_example}
\end{figure}

To model $ P(\textbf{x}|u_i) $, we represent $ u_i $ by a parametric Hidden Markov Model (HMM) $M_{u_i}$ defined by the set of states $S_{u_i} = \{ s_1, s_2, s_3, \dots , s_n \}$, the set of observations \mbox{$ X_{u_i} \subset \mathbb{R}^m$} with $ m $ as the dimension of the input sequence, the state transition probability matrix \mbox{$ A_{u_i} \in \mathbb{R}^{n \times n}$} and the observation probability matrix\\ \mbox{$ B_{u_i} \in \mathbb{R}^{n \times m} $}. In our approach, the HMMs are defined by a strict left-to-right feed forward topology, thus, only self-transition and transitions to the next state are allowed as shown in Figure~\ref{fig:HMM_graph_example}. 


Sampling from a Markov model $M_{u_i}$, produces a sequence of states $\mathcal{S} = (\textbf{S}(x_t))_{t=1,\dots,T} $ with $\textbf{S}(x_t) \in S_{u_i}$.
The joint probability that the input sequence $ \textbf{x} $ and the sequence $\mathcal{S}$ generated by the Markov Model $ M_{u_i} $ can be calculated as the product of transition probabilities  $ A_{u_i} $ and observation probabilities $ B_{u_i} $:
\begin{equation}\label{equ_unit_recog3}
\begin{split}
P(\textbf{x}, \mathcal{S}|M_{u_i}) = b_{s_n}(x_T) \prod_{t=1}^{T-1} a_{(\textbf{S}(x_t),\textbf{S}(x_{t+1}))} b_{(\textbf{S}(x_t))}(x_t) \text{ ,}
\end{split}
\end{equation}
where the transition probability from state $s_t$ to state $s_{t+1}$ is defined by  $ a_{(\textbf{S}(x_t),\textbf{S}(x_{t+1}))} \in  A_{u_i}$ and the observation probability of a state $\textbf{S}(x_t)$ is defined by a Gaussian mixture model:
\begin{equation}\label{equ_unit_recog6}
\begin{split}
b_{s}(x_t) = \sum_{k=1}^K \lambda_{ks} N(x_t; \mu_{ks}, \Sigma_{ks})  
\end{split}
\end{equation}
with
\begin{equation}\label{equ_unit_recog6b}
\begin{split}
N(x; \mu,\Sigma){=}\frac{1}{\sqrt{(2\pi)^m|\Sigma|}}\exp\left(-\frac{1}{2}(x-\mu)^T\Sigma^{-1}(x-\mu)\right) \text{ ,}
\end{split}
\end{equation}
where $ m $ is the dimension of the input sequence $ \textbf{x} $,  $ \mu $  the $m$-dimensional mean vector,  $ \Sigma $ the $m \times m$ covariance matrix and $|\Sigma|$ the determinant of $\Sigma$.

We assume that  $ P(\textbf{x}|M_{u_i})$ corresponds to $ P(\textbf{x}, \mathcal{S}|M_{u_i}) $ by choosing the sequence $ \hat{\mathcal{S}} $ that maximizes $ P(\textbf{x}, \mathcal{S}|M_{u_i}) $, i.e., 
\begin{equation}\label{equ_unit_recog4}
\begin{split}
\hat{\mathcal{S}} = \underset{\mathcal{S}}{\operatorname{argmax}}{\left( \prod_{t=1}^{T-1} a_{(\textbf{S}(x_t),\textbf{S}(x_{t+1}))} b_{(\textbf{S}(x_t))}(x_t) \right)} \text{ ,}
\end{split}
\end{equation}
and the probability follows by
\begin{equation}\label{equ_unit_recog4b}
\begin{split}
 P(\textbf{x}|M_{u_i}) =  P(\textbf{x}, \hat{\mathcal{S}}|M_{u_i})  \text{ .}
\end{split}
\end{equation}

This leads back to the idea that the model $ M_{u_i} $ is a representation of the given unit $ u_i $ and that the best path through $ M_{u_i} $ corresponds to the probability $P(\textbf{x}|u_i)$ of the observation of a unit $u_i$ given an input sequence $\textbf{x}$ 
\begin{equation}\label{equ_unit_recog5}
\begin{split}
 P(\textbf{x}|u_i) = P(\textbf{x}|M_{u_i})  \text{ .}
\end{split}
\end{equation}

The parameters $A$ and $B$ of the HMM are optimized using Baum-Welch re-estimation. 
For the decoding of HMMs, the Viterbi algorithm is used.

\subsection{Sequence model}
\label{sec_system:seq}

The recognition of individual action units may be thought of as a first step towards the analysis of ongoing event but it is unusual and rather artificial to assume that everyday tasks may consist in only one single action unit. Rather everyday activities consist in meaningful sequences of action units. 
Modeling activities as sequences of action units exhibit several advantages compared to treating it as a single entity. First, breaking down a complex activity into smaller action units allows not only for the recognition of the activity as a whole, but also for the parsing of the underlying sequence into action units. Second, the bottom-up construction of activities by composition of generic action units allows for a richer representation for efficiently learning novel activities composed of action units previously learned.

\begin{figure}
\centering
\includegraphics[width=0.9\linewidth]{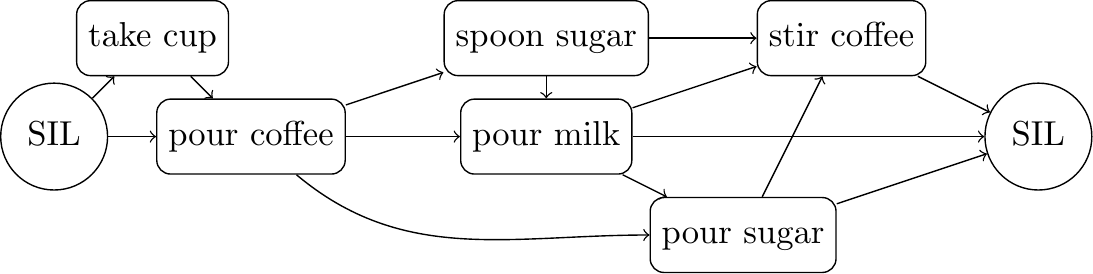}  \\ 
\caption{ Sample grammar used for the activity ``prepare coffee''. Each box represents an action unit (and thus an inidvidual HMM). ``SIL'' refers to the background (silence) class in our dataset, which is mandatory at the beginning and end of each sequence.}
\label{fig:grammar_example}
\end{figure}

\begin{figure*}[t]
\centering
 \includegraphics[width=0.9\linewidth]{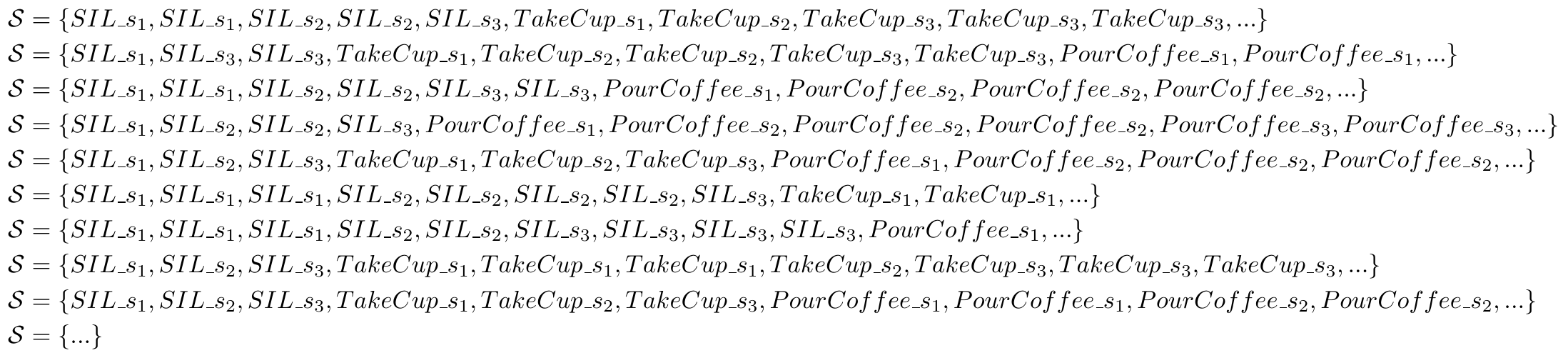}  \\ 
\caption{The nine most likely paths after processing 10 frames of an input video. Inference corresponds to finding the path with the highest probability. This includes a path through individual action units and a path through the HMM of each selected action unit. For the example shown here, it is assumed for simplicity that each action unit is represented by an HMM with three states. } 
\label{fig:result_possPaths}
\end{figure*}


We use a grammar notation based on the extended Backus-Naur form to model activities as a combination of action units. The grammar is automatically generated from the segmentation transcripts of the training data. An example is given in Figure~\ref{fig:grammar_example}. 
The recognition of sequences is based on the token passing concept for connected speech recognition~\cite{Young89tokenpassing}, augmenting the partial log probability with unit link records describing the transition from one unit to the next. To compute the most probable sequence, the Viterbi algorithm is used. At any frame $t$, the link records can be traced back to get the current most probable path, i.e., the most probable combination of units, and the position of the unit boundaries, i.e., the segmentation of the sequences until the current frame.

\section{Evaluation}
\label{sec:evaluation}

\subsection{System description}
\label{sec:systemdescription}

As features, we use dense trajectories~\cite{Wang2013Dense}. The dimensionality of the feature descriptors is first reduced from 426 dimensions to 64 dimensions by PCA, following the procedure described in~\cite{Oneata2013Action}. 
To compute the Fisher Vectors (FVs), we sample 200,000 random features to learn Gaussian mixture models. The FV representation is computed for each frame over a sliding window of size 20 frames using \emph{vlfeat}~\cite{vedaldi08vlfeat}. The dimensionality of the resulting vector is then reduced to 64 dimensions again using PCA. Thus, each frame is then represented by a 64-dimensional reduced FV. We further apply a L2-normalization to each feature dimension separately for each video clip. 

In our implementation, we use the open source Hidden Markov Toolkit HTK~\cite{Young2006Htkbook}.
For the training, inidvidual units are extracted  and one HMM is trained for each unit. The number of states of the corresponding HMM is determined relative to the mean length $\hat{T}$ of the training samples, i.e.\  $n = \frac{\hat{T}}{10}$. 
We initialize the HMM by splitting all sequences evenly over time and assigning each sub-sequence to individual states. This initialization is possible as the HMMs are built in a left-to-right order. The HMM transition probabilities are initialized such that $a_{j, j} = 0.9$ and  $a_{j, j+1} = 0.1$. 

The Viterbi algorithm is applied to find the most likely state sequence for each training sequence and the HMM parameters are updated according to the newly estimated state sequence. The process is repeated until no further increase in likelihood is gained. After initialization, the states of individual action units are re-estimated by the Forward-Backward algorithm, optimizing the joint probability of states and frame inputs.

As generative models are prone to overfitting when given imbalanced training data, we set a lower bound of a 50 samples and an upper bound of 80 samples per action unit for training. When fewer training samples are available , we generate artificial samples by minority oversampling. Random down-selection is used when more than 80 samples are available.

\begin{figure}
\centering
 \includegraphics[width=0.9\linewidth]{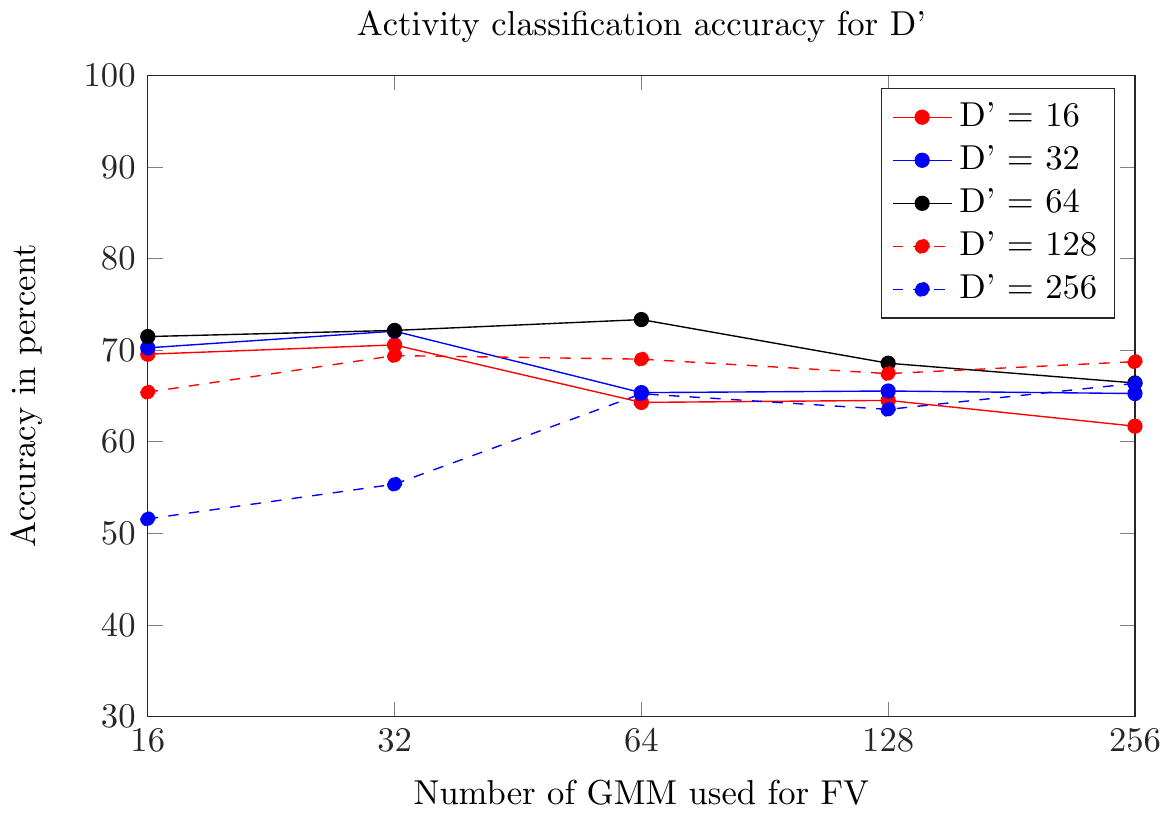}  \\
\caption{Results for activity recognition using the first D' = [16, 32, 64, 128, 256] principal components of the FV representation after PCA.  } 
\label{fig:resActivity}
\end{figure}

\subsection{Action unit classification}

We first evaluate the performance of the model for the classification of individual action units, i.e., the classification of pre-segmented videos into 48 coarse or 178 fine scale action unit classes. The task is analogous to action clip-based action classification. 
Each video segment is classified independently using \eqref{equ_unit_recog}.
We compare the classification accuracy of the HMMs with a linear SVM using the same feature representation with the exception that the FV representation is computed for the entire segment instead of each frame. In contrast to HMMs, which model the temporal relation of the observations, the SVM approach aggregates the observations independently of their temporal order.

\begin{table}[t]\scriptsize
   \centering
\begin{tabular}{|c|c c c c c c|}
\hline
\multicolumn{7}{ |c| }{Action unit classification - Breakfast - Coarse labels} \\  
\hline 
      & GMMs = & 16 &  32 &  64 & 128 &  256 \\ 
\hline 
SVM w/o PCA &      &  19.3  &  20.9   &  21.8  &  22.2   &  23.0   \\  
SVM  w PCA    & $D'=64$ &  15.2    &  16.1    &  16.6   &  16.6    &  17.8  \\  
\hline 
HMM w PCA  & $D'=64$ &   29.5 &  30.0 &  30.6  &  26.3  &  25.5    \\ 
\hline 
\end{tabular} 
\caption{Action unit classification on the Breakfast dataset based on coarse labels with 48 classes (1712 clips).}
\label{tab:unit_FV_SVM_HTK}
\end{table}

\begin{table}[t]\scriptsize
   \centering
\begin{tabular}{|c|c c c c c c|}
\hline
\multicolumn{7}{ |c| }{Action unit classification - Breakfast Fine - Fine labels} \\  
\hline 
      & GMMs = & 16 &  32 &  64 & 128 &  256 \\
SVM w/o PCA &      &  3.4 & 4.5  &  - &   - &  -  \\ 
SVM  w PCA    & $D'=64$ & 3.0   &  3.2  &  -  &  -   & -  \\  
\hline 
HMM w PCA  & $D'=64$ &   7.8 & 7.8    & 7.9   &  7.1  & 7.0   \\    
\hline 
\end{tabular} 
\caption{Action unit classification for the Breakfast Fine dataset based on fine labels with 178 classes (804 clips).}
\label{tab:unitfine_FV_SVM_HTK}
\end{table}

Table~\ref{tab:unit_FV_SVM_HTK} and \ref{tab:unitfine_FV_SVM_HTK} provide a comparison of the two approaches based on FV representations using GMMS with 16, 32, 64, 128 and 256 components. 
After PCA, the dimensionality of the FV representation is reduced to $D'=64$. We found this value to work best on the proposed dataset (see Figure \ref{fig:resActivity}).
HMMs work better with lower-dimensional features (64 components) while SVMs work better with higher-dimensional features. 
For coarse-level action unit classification, HMMs outperform  linear SVMs. When comparing the best setting for HMM and SVM, the difference in accuracy is about 7\%.    
On Breakfast Fine,  HMMs with 64 components achieve an accuracy of 24.7\% for the classification of coarse units (chance: 2.1\%), thus a drop of about 5\% compared to Breakfast. The drop can be explained by the reduced training set of Breakfast Fine compared to Breakfast cf.~Table~\ref{tab:overview_datasets}. 



The results for fine unit classification on Breakfast Fine are reported in Table~\ref{tab:unitfine_FV_SVM_HTK}. Although the overall accuracy is lower for both approaches, the results are similar to the coarse action units, i.e., the HMMs outperform the linear SVMs.

\subsection{Activity classification}
\label{subsec_activity_classification}

We evaluate the approach for activity classification on both the Breakfast and Breakfast Fine dataset. For activity classification, a complete sequence needs to be classified into one of the 10 activity classes listed in Table~\ref{tab_units_overview}. In our approach, the action units are modelled using HMMs (Section~\ref{sec_system:unit}) and the activity sequence by a grammar (Section~\ref{sec_system:seq}). For the SVM baseline, we encode the entire sequence with a single FV representation similar to the action unit classification task described above.

\begin{table}[t]\scriptsize
   \centering
\begin{tabular}{|c|c c c c c c|}
\hline 
\multicolumn{7}{ |c| }{Activity classification - Breakfast} \\ 
\hline 
      & GMMs = & 16 &  32 &  64 & 128 &  256 \\ 
\hline 
SVM w/o PCA &      & 52.0  &  52.6  &  48.7  &  39.6  &  23.2    \\ 
SVM w PCA & $D'=64$ & 42.0  &  42.5  &  42.8  &  40.3  &  41.2   \\ 
\hline 
Grammar  & $D'=64$ &  \textbf{71.5} &  \textbf{72.2}  & \textbf{73.3} &  \textbf{68.6}   &   \textbf{66.4}  \\
\hline 
\end{tabular} 
\caption{Activity classification for the Breakfast dataset. For the grammar,  action units are modeled using HMMs and PCA. 
}
\label{tab:recogact_FV_SVM_HTK}
\end{table}

\begin{table}[t]\scriptsize
   \centering
\begin{tabular}{|c|c c c c c c|}
\hline
\multicolumn{7}{ |c| }{Activity classification - Breakfast Fine} \\  
\hline 
      & GMMs = & 16 &  32 &  64 & 128 &  256 \\ 
\hline 
SVM w/o PCA &         &  43.8  &  43.2  &  47.0   &  48.8   &    48.9      \\  
SVM w PCA    & $D'=64$ &  37.8  &  32.4  &  36.1  &  34.0     &     41.0    \\  
\hline 
Grammar (coarse)   & $D'=64$ &  \textbf{61.1} &  \textbf{63.1}  & \textbf{64.5} &  \textbf{57.8}   &  \textbf{56.6} \\
\hline 
Grammar (fine)     & $D'=64$ &  \textbf{67.3} &  \textbf{68.8}  & \textbf{70.1} &  \textbf{61.5}   & \textbf{63.9}   \\ 
\hline 
\end{tabular} 
\caption{Activity classification for the Breakfast Fine dataset. For the grammar (coarse),  action units are defined using coarse labels. For the grammar (fine), action units are defined using fine labels.}
\label{tab:recogact_FV_SVM_HTKfine}
\end{table}




We report the activity recognition accuracy for both approaches on the Breakfast and Breakfast Fine dataset in Table \ref{tab:recogact_FV_SVM_HTK} and Table \ref{tab:recogact_FV_SVM_HTKfine}, respectively. On both datasets, the grammar with HMMs outperforms the SVM by about 20\%. 
For the Breakfast Fine dataset, we also assess how the level of granularity used for training the system affect accuracy.   
The comparison shown in Table \ref{tab:recogact_FV_SVM_HTKfine} shows that using 178 action units based on the fine labels instead of 48 action units based on the coarse labels improves the accuracy for activity classification. We will analyze the impact of the granularity further in Section \ref{subsec_activity_classification}. Compared to Breakfast, the overall recognition accuracy for the grammar based on coarse units decreases by about 10\%. The same effect was observed above for action unit classification.  

\begin{figure}
 \includegraphics[width=0.9\linewidth]{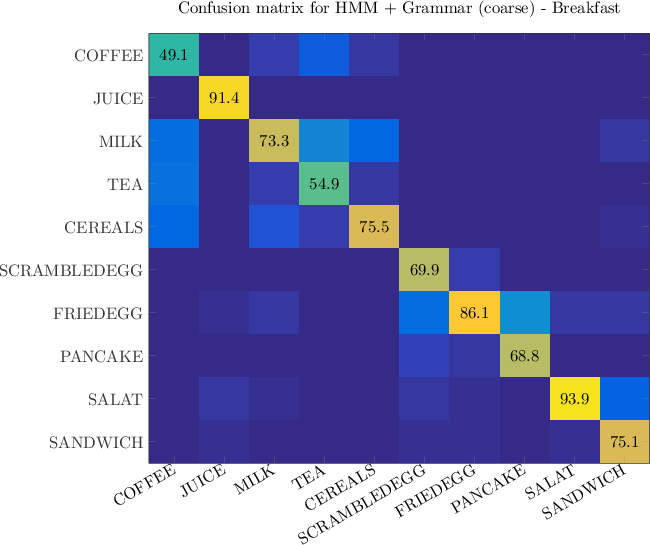}   \\ \vspace{0.2cm}
 \includegraphics[width=0.9\linewidth]{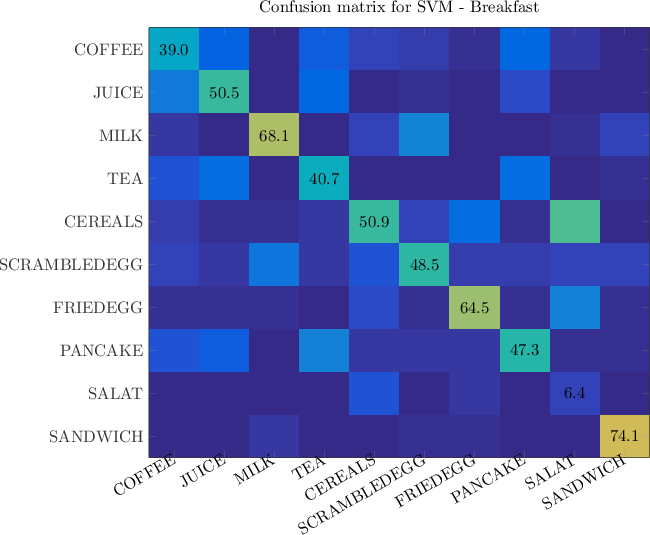}  \\ \vspace{0.2cm}
\caption{Confusion matrix for the grammar (left) and the SVM (right) for Breakfast. The grammar gets mainly confused by semantically similar activities. For instance, the preparation of various drinks (coffee, milk, tea) are confused among each other. The confusion matrix for the SVM approach does not show a clear pattern.}
\label{fig:confMat_act_coarse}
\end{figure}

\begin{figure}
 \includegraphics[width=0.9\linewidth]{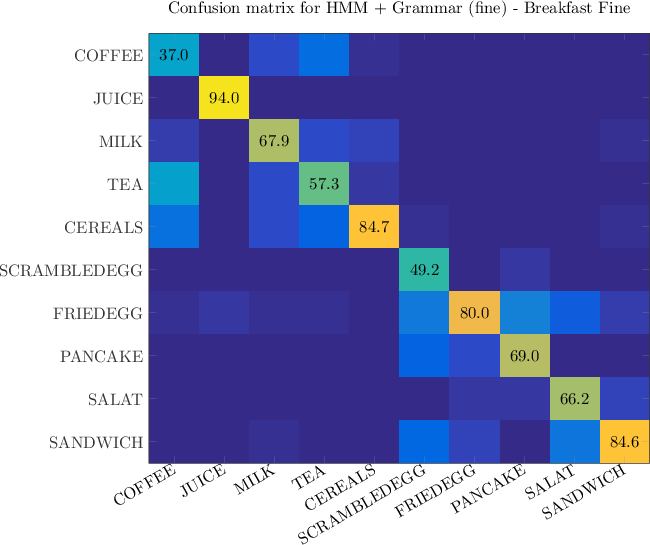}   \\ \vspace{0.2cm}
 \includegraphics[width=0.9\linewidth]{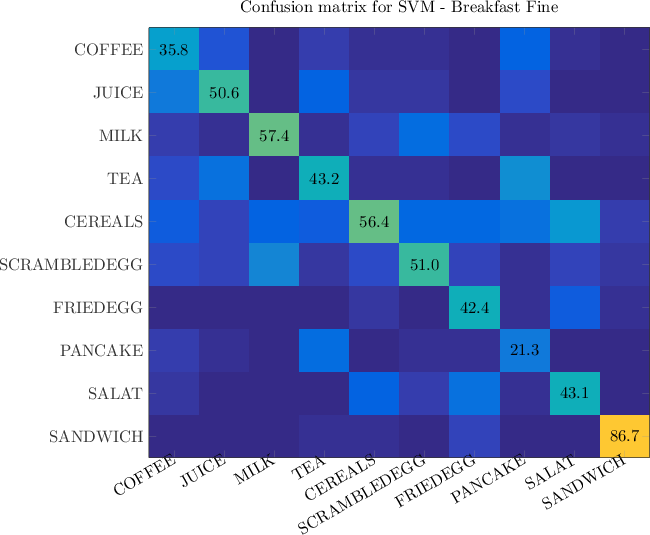}  \\ \vspace{0.2cm}
\caption{
Confusion matrix for the grammar with fine action units (left) and the SVM (right) for Breakfast Fine with 804 clips. As in the large dataset HMM based recognition groups semantically similar activities. }
\label{fig:confMat_act_coarse_fine}
\end{figure}

Figure~\ref{fig:confMat_act_coarse} and \ref{fig:confMat_act_coarse_fine} show the confusion matrices for both datasets. The grammar with HMMs, in contrast to the SVM, tends to confuse semantically-similar activities such as preparing coffee, tea or chocolate milk compared to semantically-dissimilar ones such as preparing a sandwich. Overall, the confusion matrix shows a clear grouping of activities related to the preparation of drinks (top left) vs.\ food (bottom right). The only exception is the preparation of cereals, which tends to get more confused with the preparation of drinks rather than food. When comparing the unit list in Table~\ref{tab_units_overview}, however, one can see that the preparation of cereals shares more elements with the preparation of coffee, e.g., pouring and stirring, than with the preparation of scrambled egg or other food related activities. 


\subsubsection{Granularity in activity classification}
\label{subsec_granularity}

In the previous experiments, we have observed that fine-grained action units result in a higher activity recognition rate compared to their coarse-grained counterparts. 
As can be seen from Table~\ref{tab:overview_datasets}, while there are many more classes for the fine-level action units, their mean unit length tends to be shorter (5 seconds vs.\ 26 seconds for the coarse-level action units). 

To assess the influence of the mean unit length on the overall activity recognition performance, we used the coarse labels of the Breakfast Fine dataset and split them evenly into 3, 5, 10, 15, and 20 parts resulting in a multiple for the original 48 classes, and thus reducing the mean length of the units. For 5 splits, the number of action units is artificially increased to 240 and the mean length of each unit is reduced to $5.2$ seconds.      
Table~\ref{tab:com_splits} compares the subdivided coarse units with the original coarse units and the fine units. The overall activity recognition increases with the level of granularity and almost reaches the performance of the recognition based on the fine-grained unit annotation. This implies that the approach works better with finer units. 

\begin{table}[t]\scriptsize
   \centering
\begin{tabular}{|c c c c c c|c|c|}
\hline
\multicolumn{8}{ |c| }{Granularity for activity recognition} \\  
\hline 
Splits: & 3  & 5  & 10 & 15  & 20  &  coarse &  fine \\
\hline 
 & 67.3 &  68.7  &  68.3 & 69.9  & 68.9  &   64.5  &  70.1 \\    
\hline 
\end{tabular} 
\caption{
Activity classification for Breakfast Fine. The coarse units are artificially split into smaller units.  
}
\label{tab:com_splits}
\end{table}

\subsection{Segmentation}

The third task 
is the segmentation of long sequences, i.e., the detection of action units as they appear in an unknown sequence including the start and end frames of each segment corresponding to one unit. The proposed approach based on a grammar and HMMs directly predicts the action unit and the state of an action unit for each frame as illustrated in Figure~\ref{fig:grammar_example}.
We evaluate the segmentation for both sets by looking at the overall number of frames that were correctly classified in terms of action units. Following the evaluation protocol of~\cite{Kuehne16}, we report mean over class (MoC) and mean over frames (MoF).

In Table \ref{tab:recog_FV_SVM_HTK_seg} and \ref{tab:recog_FV_SVM_HTK_segfine}, we report the segmentation accuracy for Breakfast and Breakfast Fine, respectively. On Breakfast, the coarse units are very well segmented with $56.3\%$ of all frames correctly classified. The evaluation includes the sequences that were wrongly classified, which are about 25\% of the sequences, cf.\ Table~\ref{tab:recogact_FV_SVM_HTK}. When only the correctly classified sequences are considered, the overall amount of correctly classified frames increases to 70.5\%. 
To asses the segmentation performance without grammar, we replace the grammar by a transition graph that allows transitions to and from any action unit without constraints. Without the grammar, the segmentation accuracy drops to $26.5\%$ correctly classified frames.   

We also evaluate the segmentation accuracy for the Breakfast Fine dataset using the fine-grained action units. While $31.3\%$ of the frames are correctly classified, the mean over class is only $12.2\%$. This shows the difficulty of segmenting fine-grained action units.    

The problem becomes visible when looking at the related examples for the fine segmentation in Figure~\ref{fig:confMat_act_coarse}. Even if the overall sequence is correctly recognized, the alignment, especially in case of short units, can be off and so, decreasing the overall frame segmentation performance. 


\begin{table}[t]\scriptsize
   \centering
\begin{tabular}{|c| c c c c c|}
\hline 
\multicolumn{6}{ |c| }{Segmentation - Breakfast} \\ 
\hline 
 GMMs    & 16 &  32 &  64 & 128 &  256 \\ 
\hline 
Grammar (MoC)   & \textbf{36.2} &  \textbf{36.9}  & \textbf{38.1} &  \textbf{34.0} & \textbf{32.7}  \\ 
\hline 
HMMs (MoC)  & 18.7  & 19.2  & 19.8  & 16.5  & 15.9 \\    
\hline 
\hline
Grammar (MoF)  & \textbf{54.2} &  \textbf{54.4}  & \textbf{56.3} &  \textbf{51.9} & \textbf{50.7} \\  
\hline 
HMMs (MoF) & 24.2  & 24.9  & 26.5  & 20.8 &  20.5 \\   
\hline 
\end{tabular} 
\caption{Segmentation results for Breakfast with 48 action units. For HMMs, the grammar is replaced by a transition graph that allows transitions to and from any action unit. MoC denotes mean over class, MoF denotes mean over frames.}
\label{tab:recog_FV_SVM_HTK_seg}
\end{table}


\begin{table}[t]\scriptsize
   \centering
\begin{tabular}{|c| c c c c c|}
\hline 
\multicolumn{6}{ |c| }{Segmentation - Breakfast Fine} \\  
\hline 
 GMMs     & 16 &  32 &  64 & 128 &  256 \\ 
\hline 
Grammar (MoC)  &  \textbf{11.2} &  \textbf{11.6}  & \textbf{12.2} &  \textbf{10.4}   & \textbf{10.5}   \\ 
\hline 
Grammar (MoF)  &  \textbf{28.7} &  \textbf{28.6}  & \textbf{31.3} &  \textbf{24.2}   & \textbf{26.4}   \\ 
\hline 
\end{tabular} 
\caption{Segmentation results for Breakfast Fine with 178 action units. MoC denotes mean over class, MoF denotes mean over frames.
}
\label{tab:recog_FV_SVM_HTK_segfine}
\end{table}

\begin{figure}
\centering
 \includegraphics[width=0.95\linewidth]{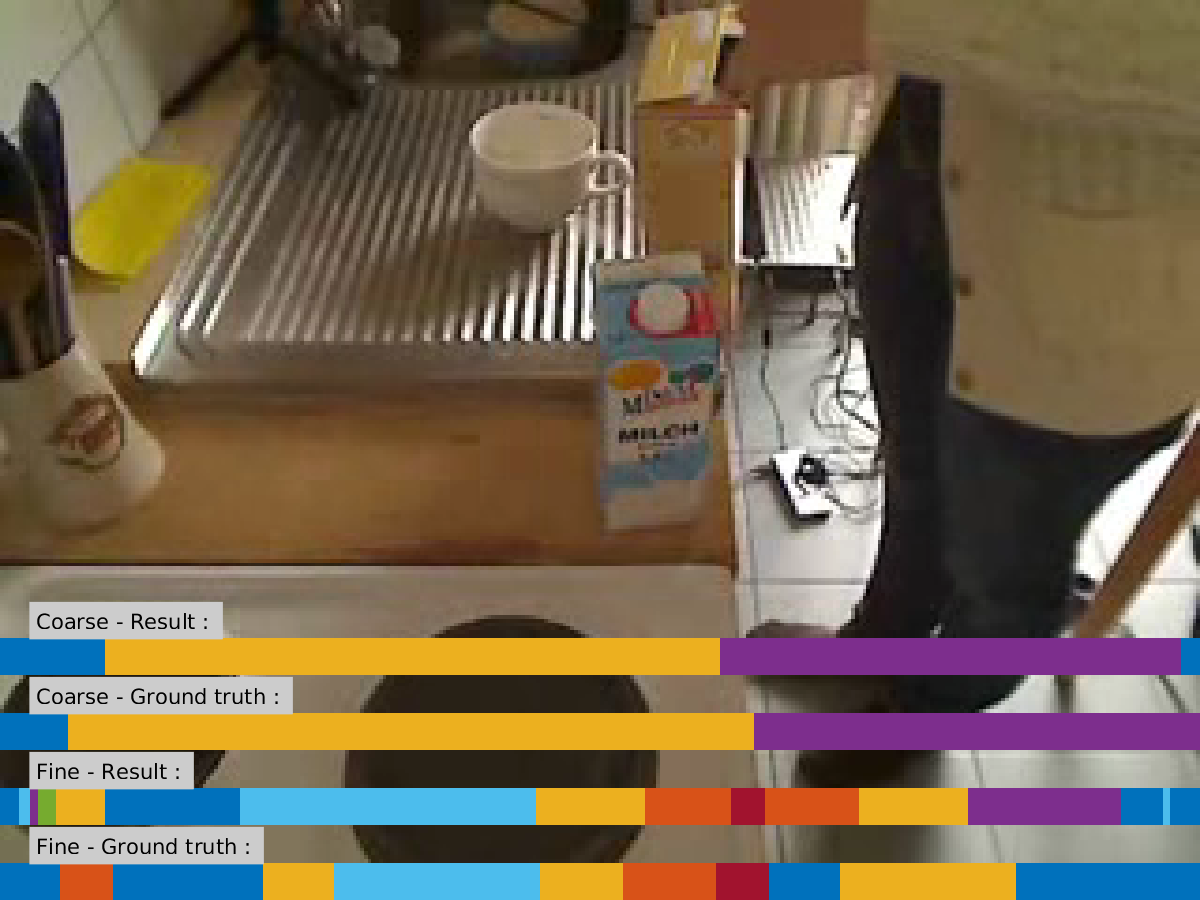}  \\ \vspace{0.2cm}
 \includegraphics[width=0.95\linewidth]{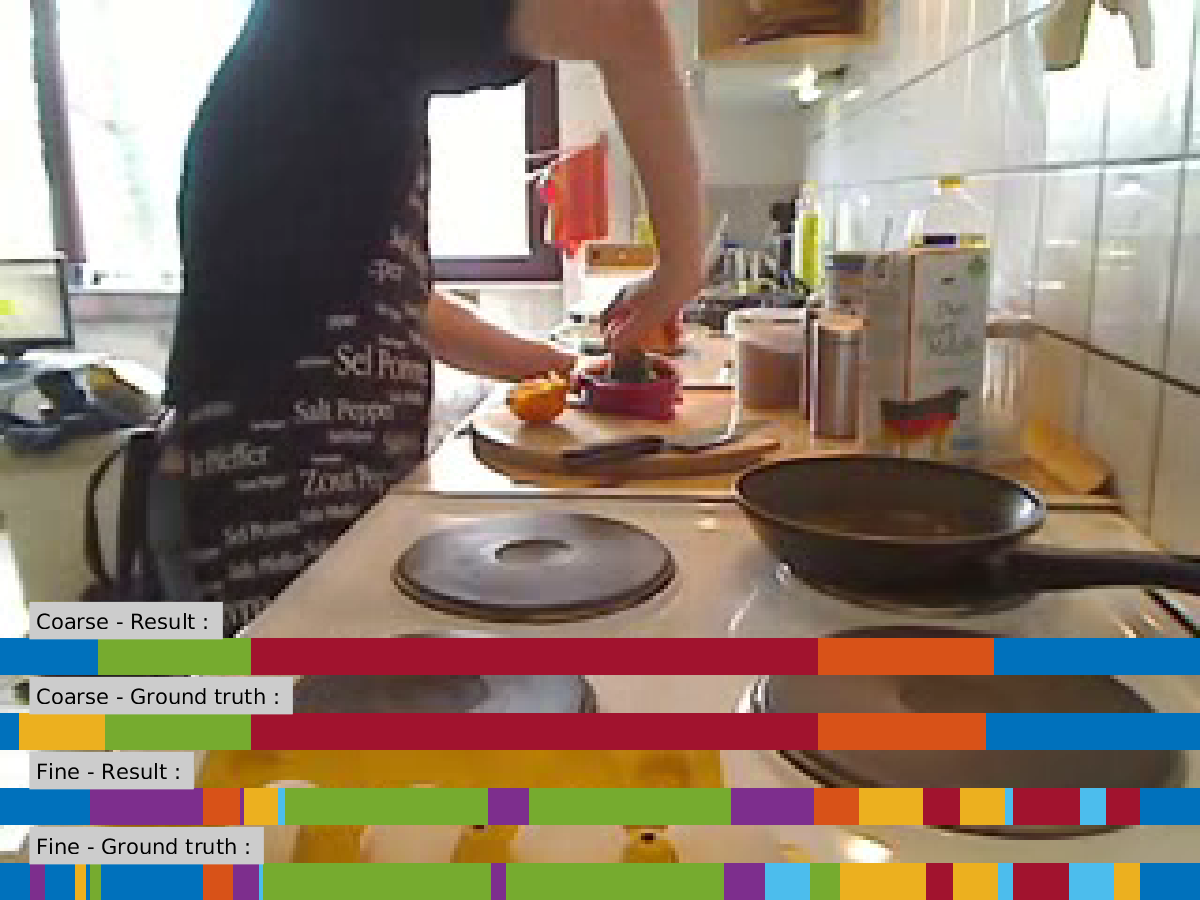}   \\ \vspace{0.2cm}
 \includegraphics[width=0.95\linewidth]{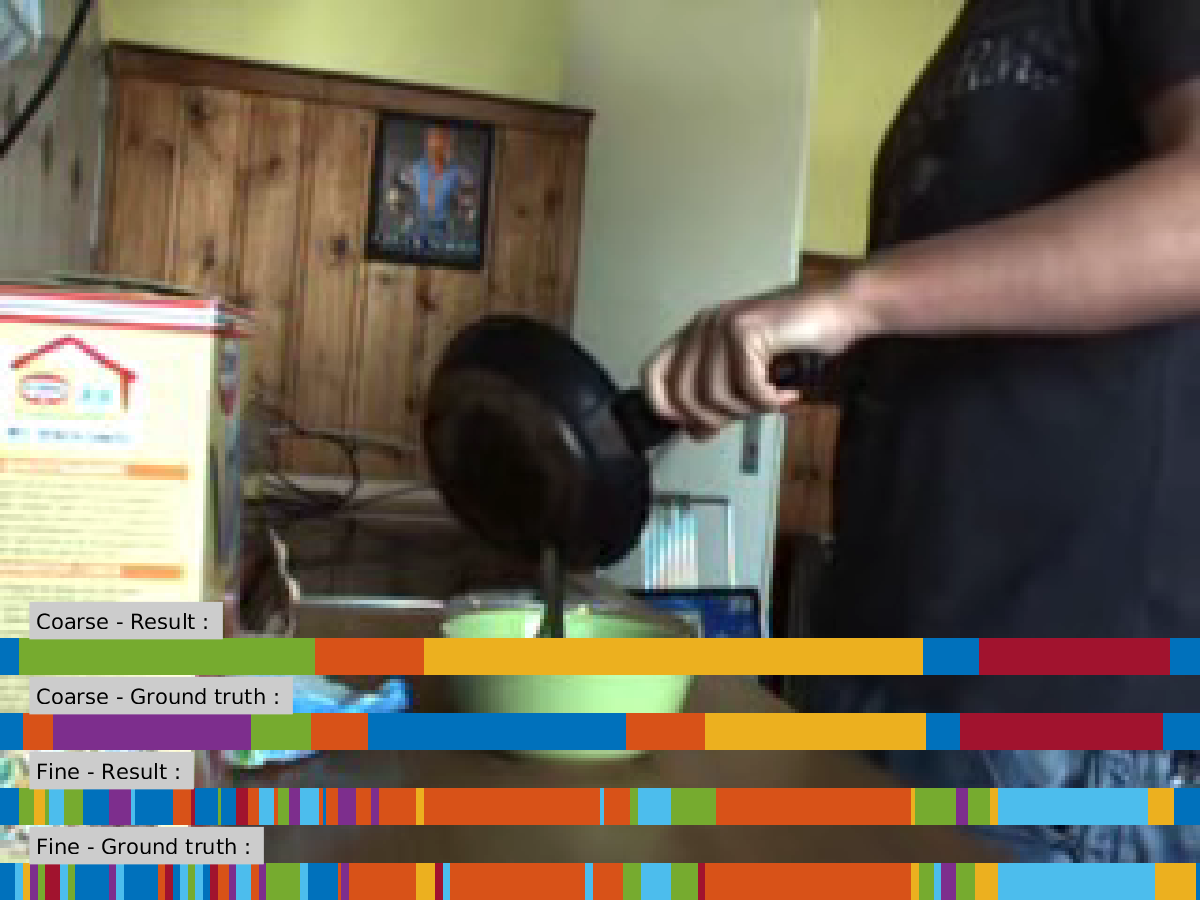}  \\ \vspace{0.2cm}
\caption{Examples of coarse and fine segmentation results for Breakfast. The upper two bar shows the recognized sequence and the respective ground truth for the coarse annotations and the lower colorbar for the fine annotation. Although the fine grained units are usually correctly classified, the overall alignment error of fine-grained units leads to a lower frame classification accuracy as in case of the coarse units.} 
\label{fig:confMat_seg_samples}
\end{figure}




\subsection{Runtime analysis}


We report the runtime for activity classification and segmentation on Breakfast Fine over all four splits without the computation of the FV representation for each frame.  
The grammar with 48 coarse action units requires $9.1$ hours for training and $1.3$ hours for testing. For 178 fine-grained action units, the training reduces to $0.86$ hours since each HMM consists of less states and is trained on shorter sequences. The inference time, however, increases to $3.85$ hours since the grammar comprises more and longer valid paths.        






\subsection{Impact of training data}

We have observed that activity recognition with coarse action units is lower on Breakfast Fine than on Breakfast due to the smaller amount of the training data.  
As the amount of training data available plays an important role in the context of generative models, we  asses how the amount of training data influences the overall recognition accuracy.
To provide a more in-depth analysis of the impact of the amount of training data, we reduce the input training data for each HMM to 50, 25, 10, 5 and 3 samples per coarse action unit on Breakfast Fine.

\begin{figure*}
\centering
 \includegraphics[width=0.4\linewidth]{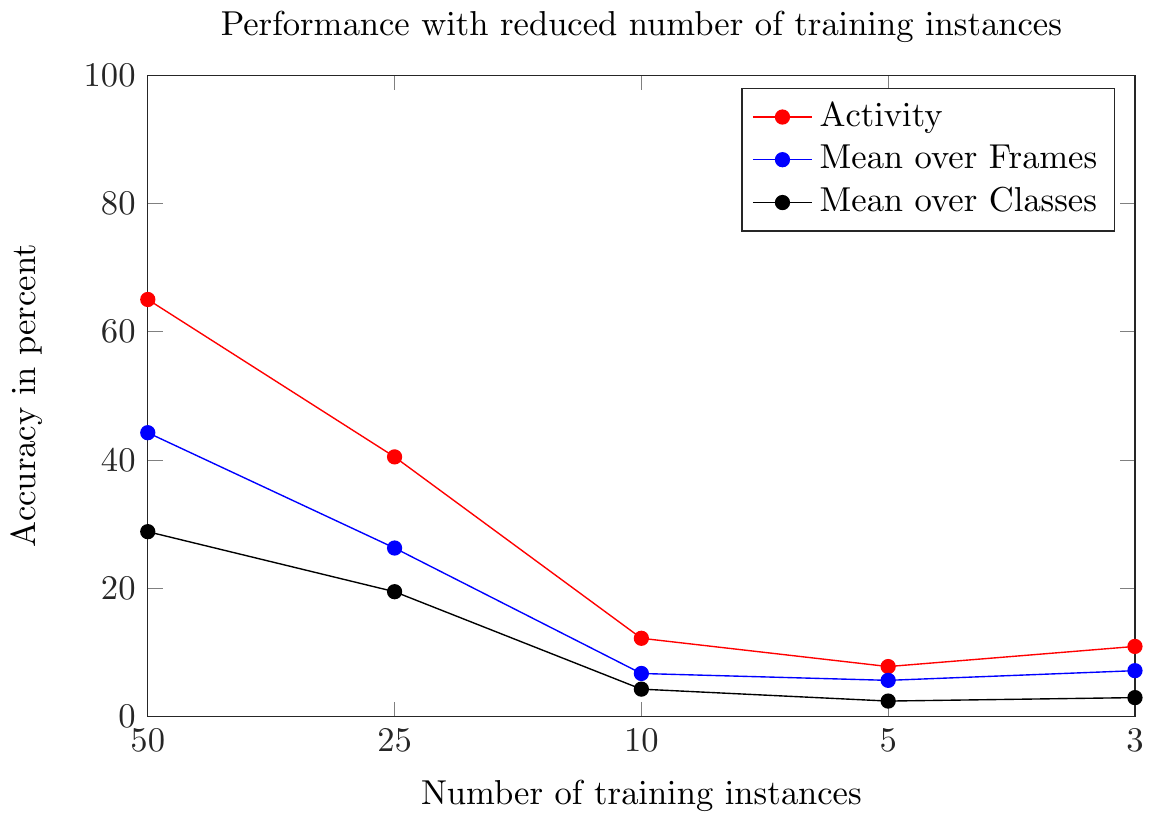}  
 \includegraphics[width=0.4\linewidth]{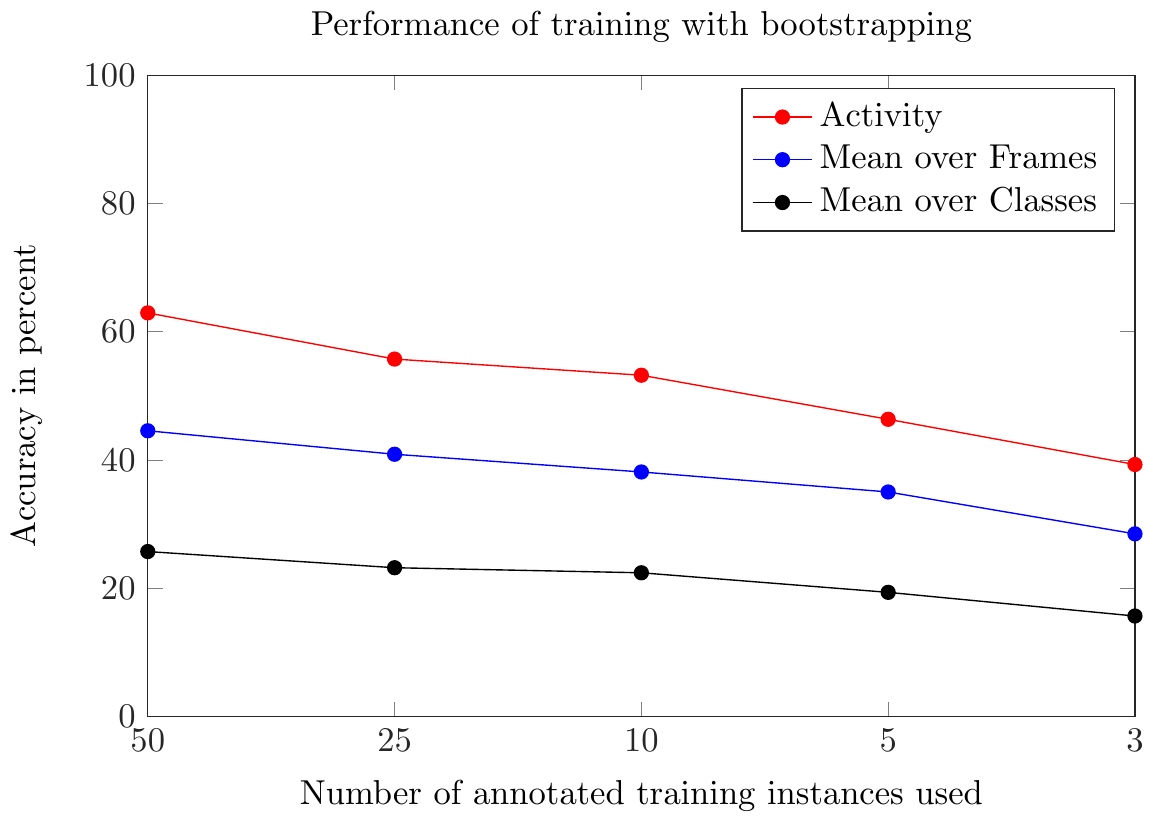}  \\
\caption{Results for activity recognition and segmentation with reduced training data without and with additional bootstrapping} 
\label{fig:resRedTraining}
\end{figure*}

Figure~\ref{fig:resRedTraining} plots the activity classification accuracy and the segmentation accuracy measured as mean over frames or mean over classes. The activity classification accuracy drops to 40\% when reducing the number of samples from 50 to 25 samples. If the number of samples per unit is less equal to 10, the activity recognition is not anymore better than chance. For the segmentation task, the same observation can be made. This shows the strong influence of the amount of training data for the approach and thus the need for large datasets like Breakfast and Breakfast Fine to study such approaches.


    
\subsection{Bootstrapping}
 
We have observed that a large amount of training data is needed to successfully train a generative model for activity classification and segmentation. The acquisition and annotation of training data, however, is very time consuming. An alternative is semi-supervised learning where only a subset is fully annotated at the frame level and most of the data is only weakly annotated with the order of action units as they appear in a video but without any alignment with the frames. The weak annotation can be considered as a transcript of the video.


For semi-supervised learning, we use bootstrapping, i.e.\ we first train the approach on the fully annotated training data as initialization and use the transcripts of the rest of the training data to reestimate the model parameters. After the model is trained on the annotated data, we infer the segmentation on the rest of the training data where the path in our grammar is given by the transcript of each video. The model parameters are then reestimated on the entire training set.  

For evaluation, we reduce the amount of fully annotated training data to  50, 25, 10, 5 and 3 samples per coarse action unit class on Breakfast Fine.  
Figure~\ref{fig:resRedTraining} plots the activity classification accuracy and the segmentation accuracy measured as mean over frames or mean over classes. The plot shows that even with a much smaller amount of fully annotated data the activity classification accuracy decreases gently to about 40\%. The same holds for the segmentation accuracy. Even though only 3 samples were used for initialization, about 30\% of all frames are correctly classified after model reestimation based on the transcribed data.




\section{Evaluation on other datasets}

We assess the performance of the proposed generative model also on four other available datasets. We first discuss its application for the action detection task on the Cha LAP dataset as an example how data augmentation in combination with majority voting can help to apply the approach even to datasets with few samples. Second, we evaluate the segmentation performance on a broad list of available datasets.

\subsection{Cha LAP dataset}

The action detection task for the Cha learning challenge differs from the so far discussed scenario since the videos contain multiple activities at the same time, i.e., the activity of each actor needs to be inferred. We therefore detect and track each actor to acquire the actor specific bounding boxes. 
The related dense trajectories are then sampled for each tracked bounding box. The dataset is very small and contains only 7 sequences for training. The segmentation quality is measured by the Jaccard index. 

\begin{table}[t]\scriptsize
   \centering
\begin{tabular}{|c|c c c c|}
\hline 
\multicolumn{5}{ |c| }{ChaLAP - Jaccard} \\  
\hline 
 GMMs  & 16 &  32 &  64 & 128 \\ 
\hline 
$D'=16$  & 0.456  & 0.492  & 0.436   & 0.339  \\ 
\hline 
$D'=32$   & 0.475  & 0.499  & 0.450   & 0.361 \\ 
\hline 
$D'=64$   & 0.416  & 0.474  & 0.418   & 0.331 \\ 
\hline
\end{tabular} 
\caption{Jaccard index for Cha LAP  }
\label{tab:ChaLAP_noMirr}
\end{table}

\begin{table}[t]\scriptsize
   \centering
\begin{tabular}{|c|c|}
\hline 
\multicolumn{2}{ |c| }{ChaLAP - Jaccard - Benchmarks} \\  
\hline
Suh~\cite{Shu15Action} & 0.4226 \\  
\hline 
Pei~\cite{Pei15Mixture} & 0.5011 \\  
\hline 
Peng~\cite{Peng15Action} & 0.5071 \\  
\hline 
 Wang\cite{Wang15Exploring}  &  0.5385 \\  
\hline 
\hline 
proposed  &  0.5239 \\  
\hline
\end{tabular} 
\caption{Jaccard index for Cha LAP  }
\label{tab:ChaLAP_benchmarks}
\end{table}

In Table~\ref{tab:ChaLAP_noMirr}, we report the segmentation accuracy for 16, 32, 64 and 128 GMM components and reduced feature dimensionality of 16, 32 and 64. The Jaccard index ranges form 0.33 to 0.5 for the different parameter combinations.  
Due to the small size of the dataset, it is difficult to determine the optimal parameters. We therefore propose two approaches to compensate for the lack of training data.        
First, we augment the training data by mirroring the videos. Averaged over all parameter settings, the accuracy increases from 0.429 to 0.452. Second, we segment the tracked bounding boxes with all parameter settings and assign a label for each frame by majority vote.  

\begin{figure}
\centering
 \includegraphics[width=0.95\linewidth]{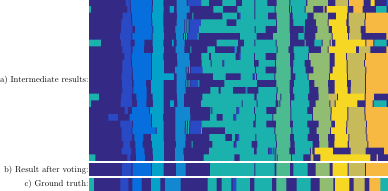}  \\ \vspace{0.2cm}
\caption{Example of a segmentation result for one actor on Cha LAP. a) shows the 24 segmentation hypotheses, b) shows the segmentation based on majority voting and c) shows the ground truth annotation. }
\label{fig:ChaLAP_voting}
\end{figure}

As shown in Table~\ref{tab:ChaLAP_noMirr}, we have 12 different parameter combinations resulting in 12 segmentation hypotheses of the original video. As for the training data, we also mirror the test video, which results in additional 12 segmentation hypotheses.     
An example for the 24 segmentation hypotheses for one actor is shown in Figure~\ref{fig:ChaLAP_voting}. The final segmentation is obtained by majority voting, i.e.\ selecting for each frame the class label that occurred the most often in all hypotheses. 

Due to the voting, we do not have to choose a particular parameter setting and achieve a Jaccard index of 0.5239. As shown in Table~\ref{tab:ChaLAP_benchmarks}, our voting procedure reaches state-of-the-art accuracy for this task despite of the small amount of training data.

\subsection{Other datasets}

\begin{figure*}[t!]
\centering
a)\includegraphics[width=0.2\textwidth]{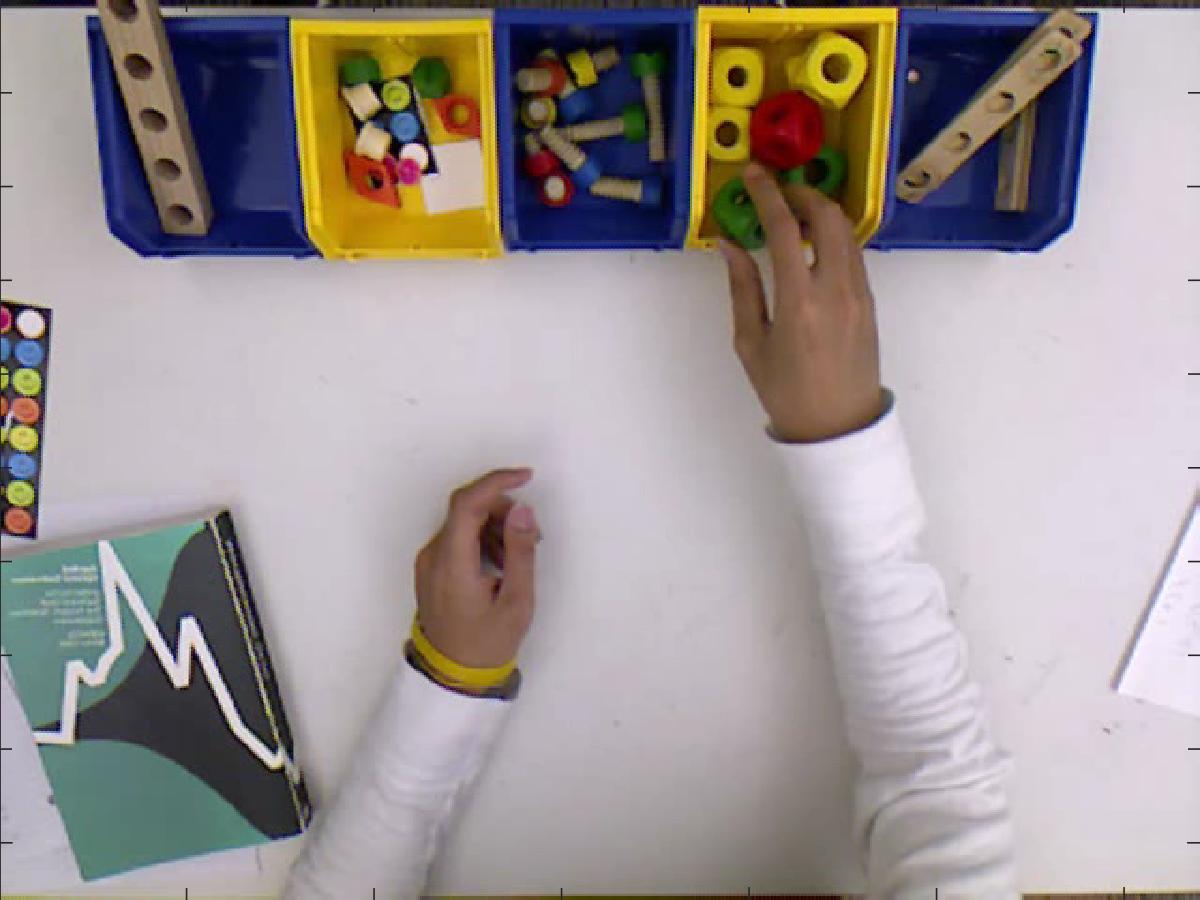}
b)\includegraphics[width=0.2\textwidth]{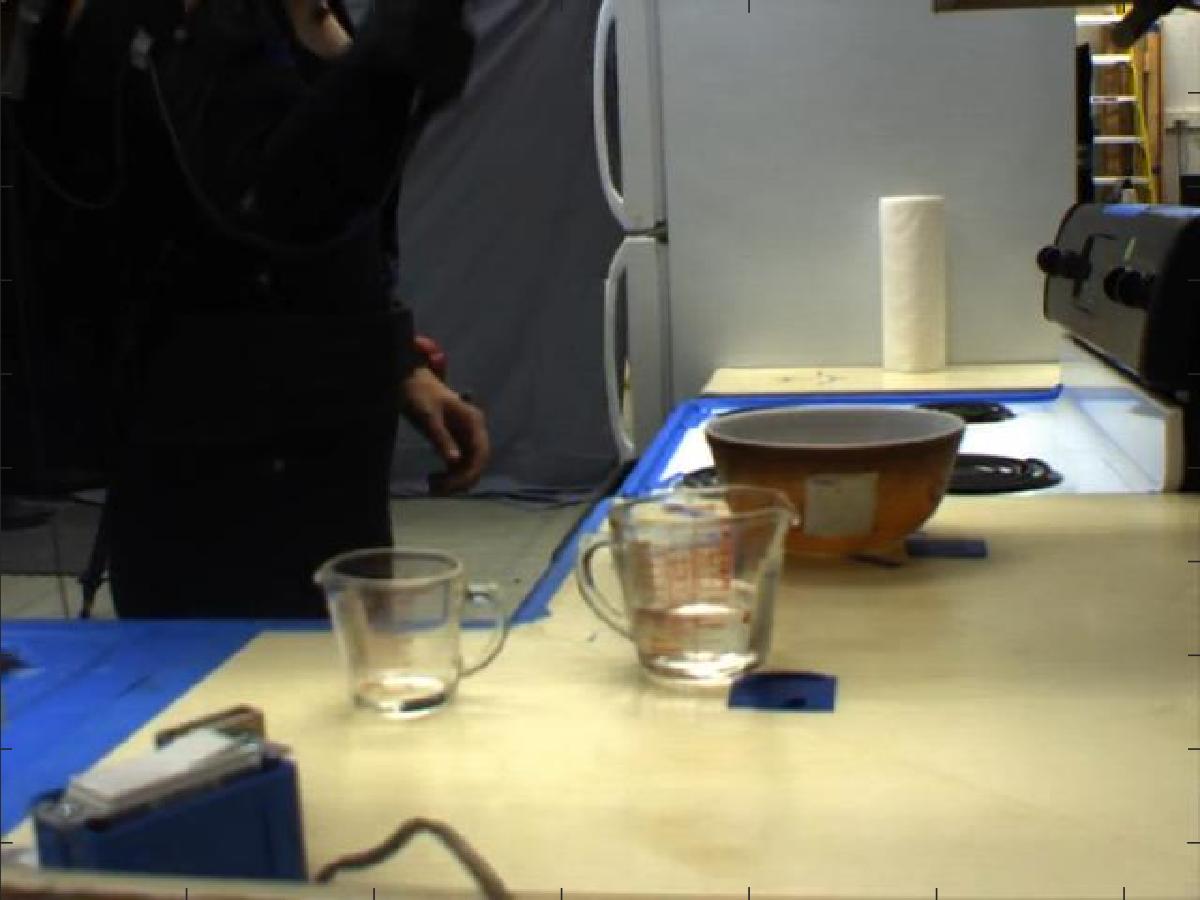}
c)\includegraphics[width=0.2\textwidth]{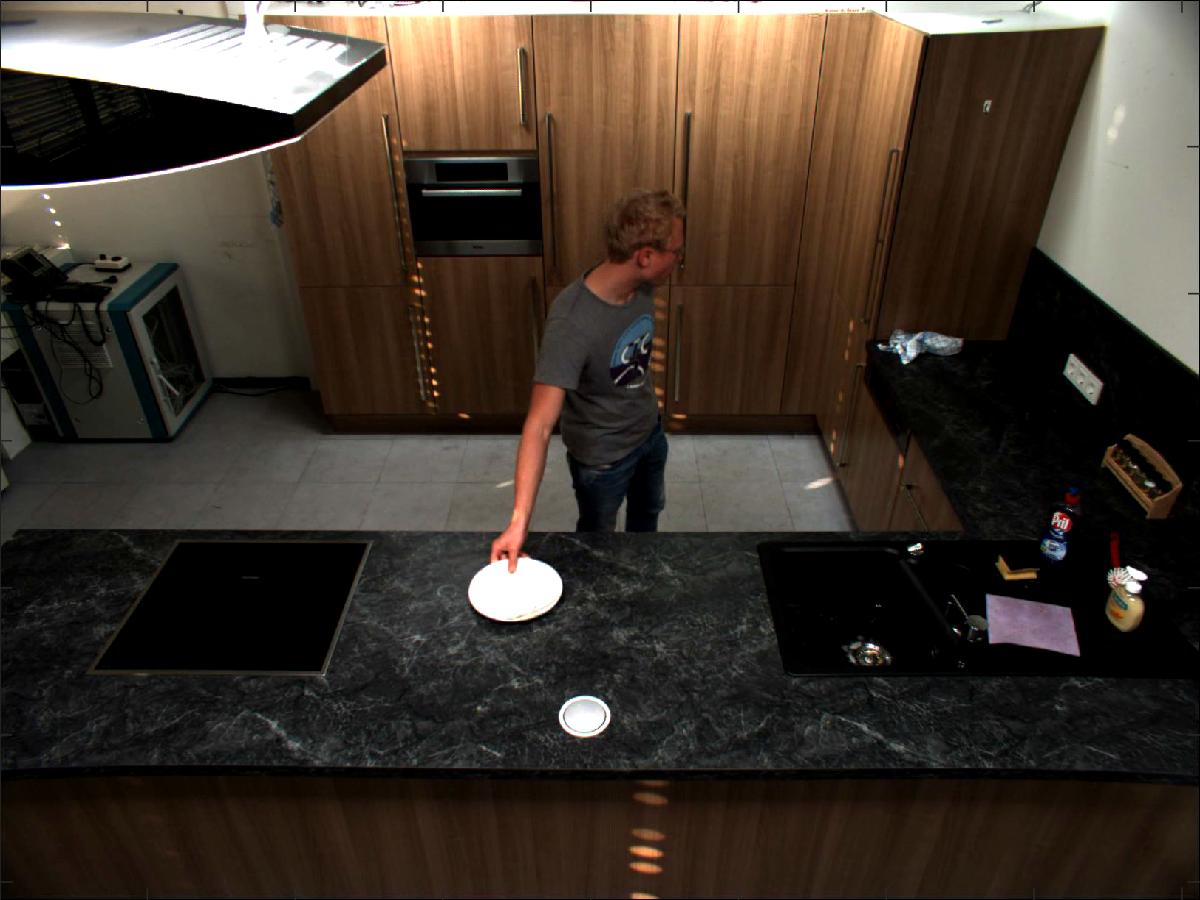}
d)\includegraphics[width=0.2\textwidth]{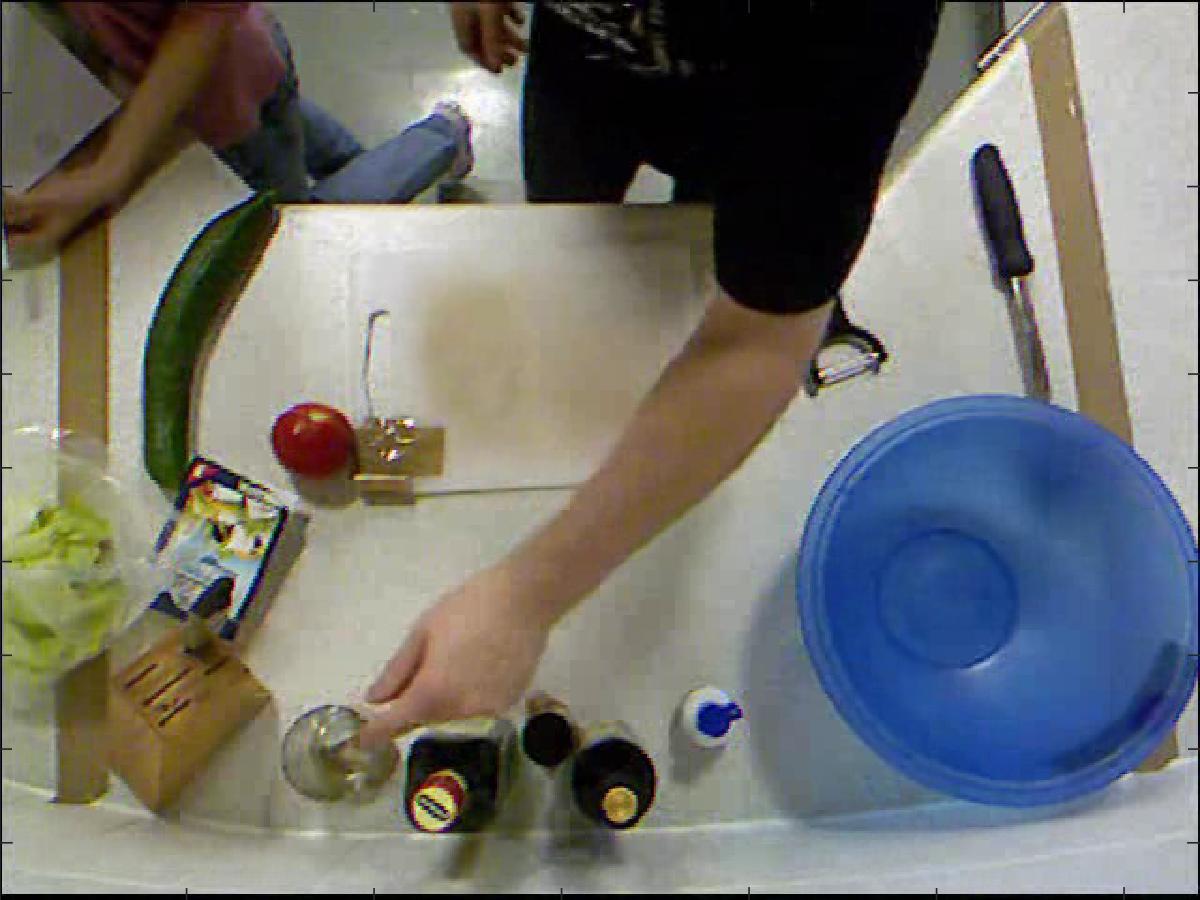}
\caption{Sample frames from the datasets used for performance evaluation: a) Toy assembly~\cite{Vo2014Stochastic}, b) CMU MMAC~\cite{Spriggs2009TemporalSegmentation}, c) MPII Cooking~\cite{rohrbach12cvprOnline}, d) 50 Salads~\cite{Stein2013Combining}.}
\label{fig:dataset_samples}
\end{figure*}

We also evaluate the parsing and segmentation performance of the proposed generative recognition approach on other available complex activity datasets that are labeled at one or more levels of granularity as listed in Table~\ref{tab:overview_datasets}. The datasets used for this evaluation are Toy assembly~\cite{Vo2014Stochastic}, CMU MMAC~\cite{Spriggs2009TemporalSegmentation}, MPII Cooking~\cite{Rohrbach2012database} and 50 Salads~\cite{Stein2013Combining}. Sample frames for each of these datasets are shown in Figure~\ref{fig:dataset_samples}. Since we observed that the accuracy of the proposed method strongly depends on the amount of training data, we report the number of training samples per class in Table~\ref{tab:overview_datasets_samples}. 
Depending on the benchmark, different measures have been proposed. For all datasets, we report the segmentation accuracy as mean over class (MoC). In addition, we report the accuracy according to the measure that was originally proposed by the corresponding dataset.


\begin{table}[t!]
   \centering
\begin{tabular}{|c|c|}
\hline 
               & Train samples used per class \\ 
\hline 
 Toy assembly          	 	 & 15-20 samples \\ 
 CMU MMAC         	  	 & 30-40 samples   \\ 
 MPII Cooking        	   	 & 12-30 samples    \\ 
 50 Salads    	 	 & 30-35 samples    \\
\hline 
\end{tabular} 
\caption{Number of used training samples per action unit }
\label{tab:overview_datasets_samples}
\end{table} 

\begin{table*}[t] 
   \centering
\begin{tabular}{|c|l|l|l|l|l|}
\hline 
\multicolumn{6}{ |c| }{Segmentation } \\ 
\hline 
GMM=        & Toy assembly        & CMU MMAC       & MPII Cooking    & 50 Salads      & Breakfast                       \\ 
\hline 
16            & 50.3 / \textit{64.3}    & 53.8 / \textit{60.8}     & 46.5 / \textit{58.5}          & 81.6           & 36.2 / \textit{54.2}      \\ 
32            & 48.6 / \textit{63.1}    & 53.7 / \textit{60.7}     & 53.9 / \textit{68.5}          & 80.4             & 36.9 / \textit{54.4}  \\ 
64           & 56.7 / \textit{67.5}    & 53.0 / \textit{60.3}     & 51.6 / \textit{63.9}          & \textbf{83.8}  & \textbf{38.1 / \textit{56.3}}   \\ 
128          & 60.5 / \textit{70.8}    & 52.5 / \textit{60.4}     & 53.9 / \textit{66.8}          & 82.0           & 34.0 / \textit{51.2}           \\ 
256          & 63.5 / \textit{72.2}    & \textbf{58.8 / \textit{67.1}} & \textbf{57.3 / \textit{71.7}} & \textbf{83.8}  & 32.7 / \textit{50.7}       \\ 
\hline 
Best    & \hspace{0.2cm}--\hspace{0.2cm} / \textbf{\textit{91.0}}~\cite{Vo2014Stochastic} & \hspace{0.2cm}--\hspace{0.2cm} / \textit{59.0}~\cite{Vo2014Stochastic} &  \hspace{0.2cm}--\hspace{0.2cm} / \textit{54.3}~\cite{Ni2014Multiple}    & 67.6~\cite{Stein2013Combining}     & \hspace{0.2cm}--\hspace{0.2cm} / \textit{28.8}~\cite{Kuehne2014Language}   \\ 
\hline 
\end{tabular} 
\caption{Overview of the segmentation results for all datasets. Accuracy is computed as the mean over all classes. For comparison, we also report the frame-based accuracy (\textit{italic}) for the Toy assembly and CMU MMAC and midpoint hit accuracy (\textit{also italic}) for the MPII Cooking dataset as used by the authors in the original studies.}
\label{tab:eval_dataset_segmentation}
\end{table*}


We compare our approach to the best reported results on each benchmark in Table~\ref{tab:eval_dataset_segmentation}. It shows that the proposed approach underperforms the best segmentation results in case of the small Toy assembly dataset. For the other datasets, which contain at least 4 hours of video, our approach significantly outperforms the state-of-the-art in terms of segmentation accuracy. 

\section{Conclusion}

We proposed a challenging large-scale dataset for human activity recognition in the wild as well as a generative approach for activity recognition and segmentation. The dataset offers the opportunity to evaluate the performance of approaches for activity classification as well as segmentation on realistic, large-scale data. The labeling at two levels of temporal granularity further allows to investigate the impact of granularity. 
On this dataset, we thoroughly evaluated the proposed generative approach that models activities by a grammar and action units as Hidden Markov Models in combination with a compact video representation based on Fisher Vectors. The experimental evaluation showed that the approach outperforms the  state-of-the-art and that generative approaches can provide high quality activity classification and segmentation results, but they need sufficient training data. While we also discussed approaches to overcome the lack of annotated training data, the two Breakfast datasets will allow to study temporally structured models more in detail.



\bibliographystyle{spmpsci}      
\bibliography{bibliography}   

\end{document}